\documentclass{article}

\usepackage{times}  
\usepackage{helvet}  
\usepackage{courier}  
\usepackage{url}  
\usepackage{graphicx}  
\usepackage{mathtools}
\usepackage{booktabs}
\usepackage{amsmath,amssymb,amsfonts}

\usepackage{caption}
\usepackage{graphicx}
\usepackage{subfigure}
\usepackage{amsthm,dsfont}
\usepackage{algorithmic}
\usepackage[ruled, noend]{algorithm2e}
\newcommand{\ignore}[1]{}
\usepackage{color}
\usepackage{authblk}
\usepackage{graphicx}
\usepackage{float} 
\usepackage{multirow}
\usepackage{amsfonts}
\usepackage{comment}
\usepackage{fancyhdr}
\usepackage{rotating}

\begin{document}
	
	\title{Heuristic Strategies for Solving Complex Interacting Large-Scale Stockpile Blending Problems}

\author{Yue Xie}
\author{Aneta Neumann}
\author{Frank Neumann}
\affil{Optimisation and Logistics, School of Computer Science,\\ The University of Adelaide, Adelaide, Australia}
\renewcommand\Authands{ and }
\maketitle
\begin{abstract}
The Stockpile blending problem is an important component of mine production scheduling, where stockpiles are used to store and blend raw material. The goal of blending material from stockpiles is to create parcels of concentrate which contain optimal metal grades based on the material available. The volume of material that each stockpile provides to a given parcel is dependent on a set of mine schedule conditions and customer demands. Therefore, the problem can be formulated as a continuous optimization problem. In the real-world application, there are several constraints required to guarantee parcels that meet the demand of downstream customers. It is a challenge in solving the stockpile blending problems since its scale can be very large. We introduce two repaired operators for the problems to convert the infeasible solutions into the solutions without violating the two tight constraints. Besides, we introduce a multi-component fitness function for solving the large-scale stockpile blending problem which can maximize the volume of metal over the plan and maintain the balance between stockpiles according to the usage of metal. Furthermore, we investigate the well-known approach in this paper, which is used to solve optimization problems over continuous space, namely the differential evolution (DE) algorithm. The experimental results show that the DE algorithm combined with two proposed duration repair methods is significantly better in terms of the values of results than the results on real-world instances for both one-month problems and large-scale problems.
\end{abstract}

\section{Introduction}

Recently, evolutionary algorithms (EAs) have been applied to many combinatorial optimization problems and proven to be very successful in real-world applications \cite{HORNG20123603,Hui2020,chiong2012variants}. Differential evolution (DE) is an efficient heuristic optimization algorithm that facilitates a population-based search in continuous multidimensional spaces \cite{neri2010recent,pham2011comparative}. The DE algorithm was first proposed in \cite{price1996differential,storn1996usage,storn1997differential}, after then, DE and its variants have been successfully applied to solve numerous real-world problems from diverse domains of science and engineering \cite{das2010differential,neri2010recent}. Recently, survey papers \cite{opara2019differential,das2016recent,das2010differential} provide an up-to-date view on DE algorithm and discuss its various modifications, improvements and uses.    

The open-pit mine production scheduling (OPMPS) problem has received a great deal of attention in recent years, both in the academic literature and in the mining industry \cite{Eduardo2017,Lamghari2012,sotoudeh2020production}. The OPMPS problem is a decision problem that seeks to maximize net present value by determining the extraction time of each block of ore and the destination to which this block is sent. The stockpile blending problem plays a significant role in OPMPS which determines the rate and quality of production involving large cash flows, as well as the stockpile blending problem takes mining scheduling upstream and process limitation and customer requirements downstream into account. Stockpile blending problem is a decision problem involving how many volumes of ore, within the stockpiles limit, should be claimed in each period, and for which parcel the ore should be sent, to maximize the volume of the valuable materials subject to the constraints that: (1) processing consume limited resources and affect the production profile in each period; (2) demands of downstream customers.  

In this work, we consider the stockpile blending problem that has been examined by Xie et al.\cite{xie2021heuristic}. In their study, the authors have considered the problem with the uncertainty in the geologic input data and applied Chebyshev's inequality to estimate the probability of constraint confidence. They have introduced two approaches to tackle the tight constraints that convert infeasible solutions into solutions without violating two complexity constraints. However, due to the complexity of the problem, they only considered a one-month stockpile blending problem with a reduced model and investigated their approach with the instances they created. 

To improve the research on the stockpile blending problem, we focus on the large-scale stockpile blending problem. A challenge in solving a stockpile blending problem is that its scale can be very large since there are maybe a large number of stockpiles and many parcels under planned, which is also a challenge to OPMOS. To the best of our knowledge, the large-scale stockpile blending problem has never been studied independently, although it is important for real-world mining engineering. In this paper, we introduce a realistic model of the stockpile blending problem containing a large-scale plan, and we describe the related input parameters of production processes in the real-world situation.
Moreover, we introduce an approach based on the DE algorithm for the large-scale problem and investigate the performance of the approach by examining real-data instances. 

\subsection{Related Work}

The task of the OPMPS problem is to generate a plan to guide the sequence of mining blocks of the ore body that ensure delivery of the tonnes and grade of the mineral raw material to the mill in the period under consideration. Initially, the OPMPS problem is first described by Johnson et al. \cite{johnson1968optimum} as a mixed integer linear model without considering a stockpile and led to the research that formulated the OPMPS as a Mixed-Integer Program (MIP) with binary variables \cite{johnson1968optimum,bley2012solving,Topal}. To address the challenge of the large-scale OPMPS problem, Osanloo et al. \cite{Osanloo} reviewed different models and algorithms for long-term OPMPS. They discussed the advantages and disadvantages of the deterministic and uncertainty-based approaches to solving the long-term production planning problem. However, when the research on OPMPS becomes more complex and more realistic, and the problem subjects to blending resource constraints, it becomes a challenge
for the MIP to solve the problem. Lipovetzky et al. \cite{Nir2014} introduced a combined MIP for a mine planning problem, which devises a heuristic objective function in the MIP and can improve the resulting search space for the planner. Samavati et al. \cite{samavati2017local} proposed a heuristic approach that combines local branching with a new adaptive branching scheme to tackle the OPMPS problem.

In the real-world application, stockpile plays an indispensable role in OPMPS which is used to store the material with different grades and increase efficiency of the mill. Jupp et al. \cite{jupp2013role} proposed four reasons for stockpiling before processing: buffering, blending, storing, and separating material with different grades. Robinson et al. \cite{robinson2004much} came to the conclusion that blending material in the stockpile can lead to grade variation reduction.
Some papers introduced approaches to represent the open-pit mine production scheduling with stockpiling (OPMPS+S) problems as nonlinear-integer models, and assumed that the material mixing homogeneously in the stockpile, however, this problem is difficult to solve. Akaike et al. \cite{akaike1999strategic} proposed a model for mine planning considering a stockpile, however there is no blending in the stockpile and the material grade in the stockpile is the same as the block. Moreno et al. \cite{moreno2017linear}  introduced a linear integer model to consider stockpiling in OPMPS and proofs their model is better than other models by comparing the objective function values. Recently, Rezakhah et al. \cite{rezakhah2019open} used a linear-integer model to approximate the OPMPS+S problem which forces the stockpile to have an average grade above a specific limit. 

The rest of the paper is organized as follows. In the next section, we present the model of the stockpile blending problem and two repair operators for the tight constraints. Afterward, we introduce the heuristic search approaches for the one-month and large-scale stockpile blending problems. We report on our experimental results for one-month and  long-term stockpile blending, and finish with some concluding remarks.

\section{Nonlinear model of Stockpile Blending Problem}

In a real-world application, a blending strategy is affected by the corresponding mining plan which decides the volume and quality of material hauled from mine to stockpiles. The blending strategy also leads to grade-level changing in stockpiles. The created parcels have to respond to the market plan which provides the requirements, such as tonnes concentrate and the total duration of all parcels from downstream process and customers demand. 

\begin{table}[t]
\label{tab:notation}
\centering
 \scalebox{0.9}{
  \makebox[\linewidth][c]{
  \tabcolsep=1cm
\begin{tabular}{ll}
\toprule
\multicolumn{2}{c}{\textbf{Indices and sets}}  
\\
\cmidrule(l{2pt}r{2pt}){1-1}
\cmidrule(l{2pt}r{2pt}){2-2}
Name & Description  \\
\midrule
 $s\in \mathcal{S}$  & stockpiles; $1,\ldots,S$\\
 $p\in \mathcal{P}$ & parcels; $1,\ldots,P^m$\\
$m \in \mathcal{M}$ & month; $1,\ldots,M$\\
$e_p^m$ & the $p$-th parcel scheduled in month $m$ \\
$o$ & material; $\{Cu, Ag, Fe, Au, U, F, S\}$\\
\bottomrule
\end{tabular}}}
\end{table}

\begin{table}[h]
\label{tab:notation}
\centering
 \scalebox{0.9}{
  \makebox[\linewidth][c]{
  \tabcolsep=1cm
\begin{tabular}{ll}
\toprule
\multicolumn{2}{c}{\textbf{Variables}}  
\\ 
\cmidrule(l{2pt}r{2pt}){1-1}
\cmidrule(l{2pt}r{2pt}){2-2}
Name & Description \\
\midrule
      $x_{ps}^m$ & fraction of the $p$-th parcel in month $m$\\ 
       & claimed from stockpile $s$ \\
       $t_p^m$ & produce time (duration) for $p$-th parcel \\
       & in month $m$ \\
        $w^m_p$ & volume of parcel $p$ in month $m$\\
         $\theta^m_{ps}$ & tonnage stores in stockpile $s$ after\\
           &  providing material to parcel $p$ in month $m$\\
    $c^m_p$ & copper tonne in $p$-th parcel in month $m$ \\
    $g^{om}_p$ & grade of material $o$ in $p$-th parcel in month $m$\\
    $ \tilde{g}^{om}_{ps}$ & grade of material $o$ in stockpile $s$ \\ 
     & when proving parcel $p$ in month $m$\\
    $k^m_p$ & tonne concentrate of parcel $p$ in month $m$\\
    $r^{Cu,m}_p$ &  Cu recovery of parcel $p$ in month $m$\\
    $r^{F,m}_p$ &  F recovery of parcel $p$ in month $m$\\
    $r^{U,m}_p$ & U recovery of parcel $p$ in month $m$ \\
\bottomrule
\end{tabular}}}
\end{table}

\begin{table}[h]
\label{tab:notation}
\centering
 \scalebox{0.9}{
  \makebox[\linewidth][c]{
  \tabcolsep=1cm
\begin{tabular}{ll}
\toprule
\multicolumn{2}{c}{\textbf{Parameters}}  
\\ 
\cmidrule(l{2pt}r{2pt}){1-1}
\cmidrule(l{2pt}r{2pt}){2-2}
Name & Description \\
\midrule
 $T_p^m$  & binary parameter, if $T^m_p=1$, parcel  $p$ is the parcel \\
  & need to prepare in month $m$, if $T^m_p=0$ otherwise\\
     $\delta$ & discount factor for time period\\ 
  $\tilde{\phi}$ & factor in chemical processing stage \\
 $\phi^{Au}$ & factor of Au in chemical processing stage \\
  $\phi^{U}$ & factor of U in chemical processing stage \\
  $\phi^{Fe}$ & factor of Fe in chemical processing stage \\
  $\phi^{Cu}$ & factor of Cu in chemical processing stage \\
  $(\gamma_1,\gamma_2)$  & factor of copper percentage within the produced \\
    & copper concentrate \\
  $\mu^{Fl}$ & factor of Fl recovery \\
  $\mu^{U}$  & factor of U recovery \\
  $(\mu^{Cu}_1, \mu^{Cu}_2)$  & factor of copper recovery \\
  $D^m$   & duration of month $m$ \\
  $H^m_s$   & tonnage of material hauled to stockpile $s$ in month $m$\\
  $G^{om}_s$ & grade of material $o$ that shipping to the stockpile $s$\\ 
  & in month $m$\\
  $K^m_p$ & expected tonne concentrate of parcel $p$ in month $m$\\
  $B^{F}$  & upper bound of F recovery \\
  $B^{U}$  & upper bound of U recovery \\
  $B^{Cu}$  & lower bound of Copper grade\\
  $D^{Cu}$  & bound of the difference between copper grades\\
  &  of parcels \\
  $N_m$ & number of planning parcels in month $m$ \\
\bottomrule
\end{tabular}}}
\end{table}

We present the nonlinear model of the stockpile blending problem in this section. We only consider the problem that contains a one-month plan as a unit sub-problem of the stockpile blending problem. We first introduce notation of \textit{Indices and sets}, \textit{Variables} and \textit{Parameters}, and we provide the math. We use the term "material" to include ore, i.e., rock that contains sufficient minerals including metals that can be economically extracted and to include waste.

Now, we present the non-linear model of the problem. In \cite{xie2021heuristic}, the authors formulated the stockpile blending problem with chance constraints in a simplified version by replacing complexity processes with constant parameters. Here, to discuss the problem that matches the real-world situation and is more complex, we formulate the stockpile blending problem without losing any information and describe the production processes with their corresponding input variables and parameters.

As shown in the model, a solution $X=\{X_1,\ldots,X_M\}$ consists of $M$ vectors where vector $X_m=\{x^m_{11},\ldots, x^m_{1S},\ldots, x^m_{P^mS}\}$ denotes the decision variables of parcels in month $m$, and $x^m_{ps}$ is a continuous variable in $(0,1)$ which indicates the percantage that the volume of material provided by stockpile $s$ to $p$-th parcel in month $m$. 

The objective function (\ref{obj:function}) is the sum of the volume of copper in the production concentrate of parcels, which are inextricably intertwined with metal grades and the duration of parcels. It should be noted that the calculation methods $f_1$ to  $f_5$ are represented a series of non-linear complexity calculation process specified given by our industrial partner that we are not able to publish. 

In the model, Constraint (\ref{con:duration}) ensures the sum of duration in each month is less or equal to the available production duration. Constraint (\ref{con:variables}) forces the sum of the decision variables of a parcel to equal one. Equation (\ref{con:parcelgrade}) calculates the material grades of each parcel which are related by the material grades in stockpiles and the decision variables of the parcels. Constraint (\ref{con:cugradebound}) guarantees the copper grade of parcels should be at least a given bound. $f_2$ in function (\ref{con:parceltonne}) denotes the calculated process of the volume of parcels. $f_3$ in function \ref{con:concentrate} denotes the calculated process of the tonnes concentrate of parcels, and the value of $k_p^m$ is limited by constraint (\ref{con:concentratebound}) which is a tight constraint in the model. Function (\ref{con:Frecovery}) and (\ref{con:Urecovery}) denote the calculated process of F recovery and U recovery respectively, and constraint (\ref{con:Fbound}) and (\ref{con:Ubound}) ensure the $r^{F,m}_p$ and $r^{U,m}_p$ of parcels are less than or equal to the given bounds. Constraint (\ref{con:stockpilegrade}) and (\ref{con:stockpilestorage}) enforces material grades balance and inventory balance for stockpiles when providing material to parcels. Since the ore shipping from mine to stockpiles happened at the beginning of every month, the stockpile material grades are updated once at the beginning of a month and constantly for all parcels in this month, and $\theta_{ps}^m$ should be no-negative.

\begin{align}
 \max \sum_{m \in \mathcal{M}} \sum_{p\in \mathcal{P}} c_p^m = \sum_{m=1}^{M} \sum_{p=1}^{P^m} f_1\left(t_p^m, g^o_{p^m}\right)
\label{obj:function}
\end{align}
\begin{align}
    s.t. & \sum_{1 \leq p \leq P^m} t_p^m \leq D^m  & \forall m \in          \mathcal{M} \label{con:duration}\\
        & \sum_{s\in\mathcal{S}} x_{ps}^m=1  & \forall p \in             \mathcal{P}, \forall m \in \mathcal{M}  
            \label{con:variables}\\
         &  g^{om}_p = \sum_{s\in \mathcal{S}} x^m_{ps} \tilde{g}^{om}_{ps} & \forall p \in \mathcal{P}, \forall m \in  \mathcal{M}
            \label{con:parcelgrade} \\
        &  g^{Cu,m}_p \geq B^{Cu}  & \forall p \in \mathcal{P}, \forall m \in \mathcal{M}  
        \label{con:cugradebound} \\
        & |g^{Cu,m}_{p} - g^{Cu,m'}_{p'}| \leq D^{Cu} & \forall p, p' \in \mathcal{P}, \forall m, m' \in \mathcal{M}  \label{con:cudifferent}\\
        &  w^m_p  =  f_2(t^m_{p}, g^{om}_p)   & \forall p \in \mathcal{P}, \forall m \in \mathcal{M} 
        \label{con:parceltonne}\\
        & k^m_p = f_3(t^m_p, g^{om}_p) & \forall p \in \mathcal{P}, \forall m \in \mathcal{M}  
        \label{con:concentrate} \\
        & ( k^m_p-1)  \leq K^m_p \leq (k^m_p+1) & \forall p\in \mathcal{P}, \forall m \in \mathcal{M}  
         \label{con:concentratebound} \\
          & r^{F,m}_p = f_4(g^{F,m}_p)  & \forall p\in \mathcal{P}, \forall m \in \mathcal{M}  
        \label{con:Frecovery}\\
        & r^{F,m}_p \leq B^F  &  \forall p\in \mathcal{P}, \forall m \in \mathcal{M} 
        \label{con:Fbound}\\
        & r^{U,m}_p = f_5(g^{U,m}_p)  & \forall p\in \mathcal{P}, \forall m \in \mathcal{M}  
        \label{con:Urecovery}\\
        & r^{U,m}_p \leq B^U  &  \forall p \in \mathcal{P}, \forall m \in \mathcal{M} 
    \label{con:Ubound}
\end{align}

\begin{align}
& \tilde{g}^{om}_{ps} =
\begin{cases}
\frac{\tilde{g}^{om}_{(p-1)s} \cdot \theta^{m-1}_{(p)s}+ G^{om}_s \cdot H^m_s}{\theta^{m-1}_{(p)s}+H^m_s}  &   \textit{if $p=1$}
  \\
 \tilde{g}^{om}_{(p-1)s} &  otherwise 
\end{cases} 
\label{con:stockpilegrade}\\
&\theta^m_{ps} =
\begin{cases}
\theta^{m-1}_{(p)s}+H^m_s - x^m_{ps} \cdot w^m_p  &   \textit{if $p=1$ } \\
\theta^m_{(p-1)s}- x^m_{ps} \cdot w^m_p  &  otherwise
\label{con:stockpilestorage}
\end{cases} 
\end{align}
\begin{algorithm}[t]
\caption{Decision variables normalized approach}
\label{alg:variablesapproach}
\KwIn{Decision vector $\{x^m_{p1},\ldots,x^m_{pS}\}$}
$a = \sum_{s=1}^S x_{ps}$\;
\For {$s=1$ to $S$}{
 $x_{ps} = x_{ps}/a$\;
}
\Return the normalized decision variables.
\end{algorithm}

As shown in the problem, we find that the decision variables of each parcels is consisted by the vector $\{x^m_{p1},\ldots,x^m_{pS}\}$ and the duration $t^m_p$, and the tight constraints (\ref{con:variables}) and (\ref{con:duration}) according to the decision variables are make it hard to construct a feasible solution. We use the two approaches introduced in \cite{xie2021heuristic}, the one (cf. Algorithm \ref{alg:variablesapproach}) used to normalized decision variables according to the constraint (\ref{con:variables}), the other approach (cf. Algorithm \ref{alg:durationfix}) repair the duration of parcels to convert an infeasible solution into a solution without violating constraint (\ref{con:concentratebound}).

\begin{algorithm}[t]
\caption{Duration repair operator}
\label{alg:durationfix}
\KwIn{$\{x^m_{p1},\ldots,x^m_{pS}\}$, $i\in \{1,..,I\}$, $j\in\{1,..,J\}$; parameter $\zeta$; available duration $\mathit{D}$} \
\KwOut{ parcel duration: $\mathit{d}\in \{0,\mathit{D}\}$}\
initialization: $\underline{d}=0$, $\overline{d}=\mathit{D}$, $d\in\{0,\mathit{D}\}$ , $k=\zeta \cdot d$ \
    \While{$d \in \{0, \mathit{D}\}$ and $k \notin \{K-1,K+1\}$ }{
    \If{$k>K+1$}{
    $d := (d+\underline{d})/2$\;
    
    $k := \zeta \cdot d$\;
    
    \If{$k>K+1$}{
    $\overline{d}=d$\;
    }\Else{
    $\underline{d}=d$\;
    }
    }
    \ElseIf{$k < K-1$}{
    $d := (d+\overline{d})/2$}\;
    $k := \zeta \cdot d$\;
    \If{$k>K+1$}{
    $\overline{d} :=d$\;
    }\Else{$\underline{d} :=d$}
    }
\Return \textit{the duration corresponding to solution $X$}
\end{algorithm}

\section{Heuristic search approaches}

In this section, we present the approach, the differential evolution (DE) algorithm which is a classical heuristic algorithm used to solve optimization problems in continuous space. However, the large-scale stockpile blending problem is difficult to solve due to the significant number of constraints which always contains more than four months and leads to large-scale decision variables and hard to solved by the DE algorithm as an integrated problem. Therefore, we introduce a strategy to address this challenge in the rest of this section. 

We start by designing a fitness function that can be used in the heuristic approach. The fitness function $f$ for approaches needs to take into account all constraints. We define the fitness function of a solution $X$ as:
\begin{align}
    f(X) = \left(u(X),v(X),w(X),p(X),q(X),g(X),h(X), C(X) \right)
    \label{fit:onemonth}
\end{align}
where
\begin{align}
    & u(X)=\sum_{m\in \mathcal{M}} \sum_{p\in\mathcal{P}} \max \{\left|K_p^m-k_p^m \right|,1\} \nonumber \\
    & v(X)= \sum_{m\in \mathcal{M}} \max\{\sum_{p\in \mathcal{P}}t_p^m -D^m, 0\} \nonumber \\
    & w(X)=\sum_{m\in \mathcal{M}} \sum_{p\in\mathcal{P}} \sum_{s \in \mathcal{S}} \min \{\theta^m_{ps},0\} \nonumber \\
    & p(X)=\sum_{m\in \mathcal{M}} \sum_{p\in\mathcal{P}} \max\{r^{U,m}_p-B^U,0\} \nonumber \\
    & q(X)=\sum_{m\in \mathcal{M}} \sum_{p\in\mathcal{P}} \max\{r^{F,m}_p-B^F,0\} \nonumber \\
    & g(X)=\sum_{m\in \mathcal{M}} \sum_{p\in\mathcal{P}} \max\{B^{Cu}- g^{Cu,m}_p,0\} \nonumber \\
    & C(X)= \sum_{m\in \mathcal{M}} \sum_{p\in\mathcal{P}} c_p^m. \nonumber
\end{align}


This fitness function is similar to that presented in the paper \cite{xie2021heuristic}, but we consider the problem in a general way by containing the component $p(X)$. In this fitness function, $u(X)$, $v(X)$, $p(X)$, $q(X)$ and $g(X)$ need to be minimized while $w(X)$ and $C(X)$ maximized. We optimize $f$ in lexicographic order \cite{fishburn1974exceptional}, and the function takes into account all constraints. According to the fitness function, an infeasible solution can at least violate one of the above constraints. Then, among solutions that meet all constraints, we aim to maximize the total copper tonnes. The fitness function (\ref{fit:onemonth}) can be used in any heuristic approach in continuous search space. In this paper, we investigate the performance of the classical DE algorithm (see Algorithm \ref{alg:de}).

\begin{algorithm}[t]
\caption{Differential evolution algorithm}
Generate initial population of size $NP$ \;

\While{stopping criterion not met}{
    \For{each individual $t$ in the population}{
     Generate three random integers, $r_1, r_2, r_3 \in (1, NP)$, with $r_1 \ne r_2 \ne r_3 \ne t$ \
     Generate a random integer $m_{rand} \in (1,n)$\;
     \For{each parameter $i$ of the individual}{
     Generate mutant vector $V_t$ and trial vector $U_t$;
     }
     Replace $X_{t}$ with the $U_{t}$, if $U_{t}$ is better according to the fitness function \;
     }
          }
\Return \textit{the best solution in the final population according to the fitness function.}
\label{alg:de}
\end{algorithm}

The DE is a well-known evolutionary computation approach developed for solving the global optimization problems in continuous search space where the objective function can be nonlinear \cite{das2010differential}. The DE algorithm is usually initialized by generating a population of $NP$ individuals with size $n$ using uniformly distributed random numbers, and these individuals are evaluated according to a fitness function. The next step of the DE is to create a mutant vector for each population. For each target vector $X_t; t =1,..,NP$ in the population, generate its mutant vector $V_t$ using the mutation method. The $DE/target-to-best/1$ strategy was adopted here, where the mutant vector $V_t$ is generated as: 
\begin{equation}
    V^t_i= X^t_i+F(X^t_{best}-X^t_i) +F(X^t_{r1}-X^t_{r2}),
    \label{de:mutate}
\end{equation}
where $F \in (0,1)$ is a user defined parameter which controls the magnitude of the difference vector, $X_{r1}$ and $X_{r2}$ are vectors that are randomly selected from the population, $X^t_{best}$ denotes the best individual in the current population. $X_t$, $X_{r1}$ and $X_{r2}$ must all be distinct from each other.

Following the mutation phase, the crossover operator is applied on the population. For each mutant vector $V_{t}$, an integer $k\in \{1,..,n\}$ is randomly chosen, and a trial vector $U_t$ is generated, with:
\begin{equation}
u_{it}=\left\{
\begin{array}{rcl}
v_{it} & & \text{if}\ rand(0,1)\leq CR \ or \ i=k \\
x_{it} & & \text{otherwise}
\end{array} \right.
\label{de.crossover}
\end{equation}
where, $i=1,2,..,n$; $x_{it}$, $v_{it}$ and $u_{it}$ are the components of $X_t$, $V_t$ and $U_t$ respectively. $rand(0,1)$ is a randomly generated number, and $CR$ is the crossover parameter, and it determines how often the trial vector $U_t$ gets its component value from the mutant vector $V_t$. Thus, a trial vector $U_t$ is generated and evaluated with respect to the fitness function. The target vector $X_t$ is compared to the trial vector $U_t$ w.r.t the fitness function, and the best one is selected to the next population.

Then, the trial vector is compared with its parent vector, and the better one is passed to the next generation, so the best individual in the population is preserved. The steps of DE are repeated until a specified termination criterion is reached. 

As discussed above, for the one-month stockpile blending problem, it is easy for applying the DE algorithm to obtain results by using the fitness function (\ref{fit:onemonth}). We propose a DE-based approach (see Algorithm \ref{alg:oneMonthDE}) by combining the classical DE algorithm and the decision variables normalized operator and the duration repaired operator for one-month problems.

\begin{algorithm}[t]
\caption{DE approach for one-month problem}
\label{alg:oneMonthDE}
Initialization: initial population of size NP with applying Algorithm \ref{alg:variablesapproach} to normalize decision variables of all parcels in this month;

    \While{stopping criterion not met }{
    \For{each individual in the population}{
    Generate the trial vector by applying the DE algorithm (Alg. \ref{alg:de});
    
    Normalize the trial vector by using solution fixed operator (Alg. \ref{alg:variablesapproach});
    
    Calculate the duration parcel by parcel by applying operator (Alg. \ref{alg:durationfix});
    
    Compare the new solution to the chosen individual via fitness function;
    
    Keep the best one for next population.
    }
    }
    \Return \textit{the best solution in the final population.}
\end{algorithm}

According to Equation (\ref{con:stockpilegrade}) and (\ref{con:stockpilestorage}) the material grades and the volume of material at the beginning of each month are affected by the blending strategy of last month and the ore hauled plan. Here, we propose to optimize the large-scale stockpile blending problem month by month. 

The problem objects to maximize the sum of $c^m_p$, the copper volume of parcels, which is calculated by the function with material grades and duration of parcels as input. Regarding the characters of the DE algorithm, the DE-based approach always obtains the feasible solution which has the highest objective value and leads to the highest copper grade. Therefore, the approach preferentially blends material for those stockpiles with high copper grades. However, constraint (\ref{con:cugradebound}) requires that the maximum difference between copper grades of parcels is less than a given threshold. It becomes extremely difficult to maintain the constant copper grades among all parcels when optimize the problem month by month.

To tackle the problem of maintaining the copper grades of parcels, we introduce a second objective function for the one-month stockpile blending problem. The second objective function satisfies the predefined maximum difference of copper grades. The developed fitness function $f'(X)$ of a solution is given as:
\begin{align}
   \resizebox{1\hsize}{!}{$f'(X) = \left(u(X),v(X),w(X),p(X),q(X),g(X),h(X), (C(X), C^*(X) )\right)$}
    \label{fit:multimonth}
\end{align}
where $C^*(X)= \sum_{p\in \mathcal{P}} x_{(ps^*)}$ and $x_{(ps^*)}$ denotes the decision variables that the stockpile $s^*$ which has the highest copper grade among all stockpiles in this month provide to parcel $p$. This specific stockpile might be different in each month and is only chosen according to the copper grades. Using the last multi-component allows to cater for the consistent copper grade of our problem. In bi-objective optimization of long-term plan blending optimization problem, the goal is to maximize $C(X)$ and minimize $C^*(X)$ with satisfying all constraints. Here, we have 
$$f'(X) \succeq f'(Y) \: \text{iff} \:C(X) \geq C(Y) \land C^*(X) \leq C^*(Y)$$ for the dominance relation of bi-objective optimization for two solutions $X$ and $Y$.

Solving the large-scale stockpile blending problem is a challenge due to the significant number of constraints and its scale can be very large. For example, an instance has ten parcels and can claim material from six different stockpiles, then the search space of this instance is a sixty-dimension space for decision vectors combined with a ten-dimension space for the duration of parcels. Although DE has become a popular and effective algorithm for continuous optimization problems, most reported studies on DE are obtained using small-scale problems. It becomes difficult for the DE algorithm on solving the large-scale stockpile blending problem within an acceptable compute time. 

Here, we propose an approach (cf. Algorithm \ref{alg:longtermDE}) which optimize the problem month by month using the fitness function (\ref{fit:multimonth}). The approach treats every one-month problem as a unit-problem and obtains a set of feasible solutions for each month by using Algorithm \ref{alg:oneMonthDE}.
Therefore, the process of this approach are, (1) for every month, the approach adopts a set of feasible solutions $Z$ of last month; (2) for every solution in the set $Z$, update the parameters ($\tilde{g}^{om}_{ps}$ and $\theta^m_{ps}$) of stockpiles for this month; (3) apply Algorithm \ref{alg:oneMonthDE} to obtain a set of feasible solutions of this month and add them to a set $Z'$. (4) loop all feasible solutions of last month, and select a fixed number from the set $Z'$ randomly. The steps of this approach are repeated until all months are reached. In this paper, we set the number of the feasible solution in each month is equal to the population of the DE algorithm.

\begin{algorithm}[t]
\caption{DE approach for long-term problem}
\label{alg:longtermDE}
\DontPrintSemicolon
Initialize the feasible solution set $Z$.

\For{$m$ in $\{1,..M\}$}{
Copy all element from $Z$ to $Z'$;

Clean $Z$;
    \For{element $z$ in $Z'$}{
    
  Update $S_{ij}$ and $L^o_{ij}$ by considering ore hauled of this month and the solution of $z$;

  Apply the DE-based approach (Alg. \ref{alg:oneMonthDE}) for this month by using fitness function (\ref{fit:multimonth});
  
  Add the feasible solutions in last population to $Z$;
  }
  
}
\Return The best solution in $Z$.
\end{algorithm}

\section{Experimental Investigation} \label{sec.expr}

This section evaluates the efficiency of proposed heuristic search approaches in the short-term problem and the long-term problem. It first compares the performance between the DE-based approach and the strategy used in real-world situation on one-month instances provided by our industrial partner. Then, the results obtained by the approach \ref{alg:longtermDE} are compared to the actually used results on large-scale stockpile blending instances. In this paper, all instances are provided by our industry partner as well as the real-world results of those instances. All experiments were performed using Java of version 11.0.1 and carried out on a MacBook with a 2.3GHz Intel Core i5 CPU. 

We first estimate the performance of the DE approach on one-month instances. The setup of the experiments and the results obtained by the different approaches are summarized in Table \ref{tab:shortReulst}. For example, the instance with index $2$ contains $2$ parcels, and the first parcel can claim material from $6$ different stockpiles, and the second parcel can claim material from the $6$ stockpiles and another stockpile which available in this period. In the implementation, each approach runs for $10^5$ fitness evaluates, and the DE approach runs with $NP=10$, $F=1.2$ and $CR =0.5$. Table \ref{tab:shortReulst} reports the performance of the approach by the average, maximum, minimum and standard deviation for 30 independent runs. Column $Parcels$ refers to the number of parcels contained in the instances, column $Stockpiles$ lists the available stockpiles for each parcel. The results list in column $Org$ is the real-world results provided by our industrial partner. Specifically, we set the original solution of instances as the initial solution of the approaches. 

\begin{table}[t]
  \centering
  \renewcommand{\arraystretch}{1.3}
  \caption{Results for one-month problem}
  \scalebox{0.84}{
 \makebox[\linewidth][c]{
    \begin{tabular}{llllllll}
    \toprule
    \multicolumn{1}{l}{\textbf{Index}} & \multicolumn{1}{l}{\textbf{Parcels}} & {\textbf{Stockpiles}} & \multicolumn{1}{c}{\textbf{Org}} & \multicolumn{4}{c}{\textbf{DE-based}} \\
    \cmidrule(l{2pt}r{2pt}){1-3}
\cmidrule(l{2pt}r{2pt}){5-8}
\cmidrule(l{2pt}r{2pt}){4-4}
          &       &       &      & \multicolumn{1}{c}{Max} & \multicolumn{1}{c}{Min} & \multicolumn{1}{c}{Mean} & \multicolumn{1}{c}{Std.} \\
          \midrule
    1     & 2     & $\{6,6\}$  & 5777.62  & \textbf{6167.46} & 5815.87 & 5989.31 & 99.40 \\
    2     & 2     & $\{6,7\}$ & 4180.72  & \textbf{4383.60} & 4222.38 & 4298.08 & 36.47 \\
    3     & 2     & $\{7,7\}$  & 5371.08  & \textbf{5611.94} & 5373.62 & 5472.12 & 65.12 \\
    4     & 3     & $\{7,7,7\}$  & 5124.92 & \textbf{5251.86} & 5129.03 & 5164.69 & 26.65 \\
    5     & 2     & $\{7,7\}$ & 5484.07  & \textbf{5600.56} & 5488.43 & 5536.35 & 27.79 \\
    6     & 2     & $\{7,7\}$ & 4334.78  & \textbf{4438.87} & 4340.60 & 4376.00 & 28.86 \\
    7     & 2     &$\{7,7\}$& 5243.46  & \textbf{5351.31} & 5250.24 & 5287.49 & 26.46 \\
    8     & 3     & $\{7,7,7\}$  & 5257.26  & \textbf{5411.48} & 5265.74 & 5342.09 & 32.35 \\
    \bottomrule
    \end{tabular}}}%
  \label{tab:shortReulst}%
\end{table}%

As can be seen from Table \ref{tab:shortReulst}, the results obtained by the DE approach reported significantly better objective value than the original results among all instances. Even the minimum results of all instances have higher values than the original results, which shows that the DE-based approach guaranties better results in every time of the 30 runs. Solving the stockpile blending problem is important in that is an important component of OPMPS which determines the quality of production involving large cash flows, therefore the results of the stockpile blending problems can reach hundreds of millions of dollars. These results show that for the one-month stockpile blending problem, the DE algorithm combines with our proposed repair operators is able to achieve higher copper tonnage than the real-world results which can lead to more than hundreds of millions of dollars benefits in real-world situations.

We now consider the instances that have large scales and investigate the performance of the DE approach in the large-scale problem and compare the results with the real-world results. Table \ref{tab:longReulst} lists the maximum, minimum and average results, and standard deviations of the DE approach and the results used in real-world situation of all instances. As shown in the Table \ref{tab:longReulst}, the maximum result of each instances have higher values than the original results which indicate more than hundreds of millions of dollars profit over the plan. Moreover, even the minimum result of each instances have higher values than the original results that shows the DE-based approach can reach better solution in term of the objective value than the original result in every run. Therefore, the results obtained by the DE approach are significantly greater than the real-world results on the large-scale stockpile blending instances.

\begin{table}[t]
  \centering
  \renewcommand{\arraystretch}{1.5}
  \caption{Results for long-term problem }
  \scalebox{0.75}{
 \makebox[\linewidth][c]{
    \begin{tabular}{rrrrlllll}
    \toprule
    \multicolumn{1}{l}{\textbf{Index}} & \multicolumn{1}{l}{\textbf{Month}} & \multicolumn{1}{l}{\textbf{Parcels}} & \multicolumn{1}{l}{\textbf{Length}} & \multicolumn{1}{c}{\textbf{Org}} & \multicolumn{4}{c}{\textbf{DE-based}} \\
\cmidrule(l{2pt}r{2pt}){1-4}
\cmidrule(l{2pt}r{2pt}){6-9}
\cmidrule(l{2pt}r{2pt}){5-5}
          &       &       &       &       & \multicolumn{1}{c}{Max} & \multicolumn{1}{c}{Min} & \multicolumn{1}{c}{Mean} & \multicolumn{1}{c}{Std} \\
          \midrule
    1     & 5     & 11    & 76    & 25938.41 & \textbf{26085.17} & 25988.38 & 26008.62 & 72.70 \\
    2     & 5     & 11    & 76    & 24495.58 & \textbf{24732.95} & 24497.95 & 24584.56 & 57.08 \\
     3    & 5     & 11    & 77    & 25558.31 & \textbf{25704.27} & 25565.20 & 25653.88 & 39.61 \\
    4     & 6     & 13    & 88    & 30273.20 & \textbf{30434.64} & 30289.02 & 30328.28 & 78.76 \\
     5     & 6     & 13    & 90    & 29739.03 & \textbf{29965.60} & 29761.57 & 29850.17 & 64.47 \\
       6    & 6     & 14    & 91    & 30815.57 & \textbf{31069.36} & 30841.72 & 30926.58 & 63.11 \\
    7     & 7     & 15    & 102   & 35516.65 & \textbf{35660.19} & 35530.73 & 35566.41 & 63.19 \\
     8     & 7     & 16    & 111   & 34996.29 & \textbf{35321.51} & 35009.17 & 35113.01 & 62.81 \\
      9    & 7     & 16    & 104   & 36025.85 & \textbf{36150.97} & 36029.17 & 36077.91 & 36.23 \\
    10     & 8     & 18    & 123   & 40773.92 & \textbf{41011.07} & 40782.25 & 40852.73 & 80.31 \\
    11     & 8     & 18    & 124   & 40206.57 & \textbf{40390.00} & 40213.26 & 40253.22 & 36.84 \\
    12    & 8     & 18    & 118   & 41297.38 & \textbf{41405.50} & 41308.73 & 41343.51 & 33.02 \\
    \bottomrule
    \end{tabular}}}%
  \label{tab:longReulst}%
\end{table}%

\section{Conclusions} \label{sec.conclusion}
This paper studied the large-scale stockpile blending problem which is formulated as a non-linear continuous model. The problem subject to a set of constraints dictated by the mine schedule and the demands of downstream customers. Due to the complexity and difficulty of the large-scale stockpile blending problem, we divided the problem into several unit problems that have a one-month duration. The proposed one-month stockpiles blending problem model exploits the problem constraints according to the mine schedule conditions and the requirements, and we introduced two repaired operators which improve the efficiency of finding a feasible solution in the mentioned approaches. We introduced a multi-component fitness function to control the usage of high-quality stockpiles for the large-scale stockpile blending problem and presented an approach for large-scale stockpile blending problem which optimizes the problem month by month. This approach guarantees the quality of solutions and the balance of used material between stockpiles. In the experiment section, we first investigated the DE algorithm combined with the two repaired operators for one-month stockpiles blending problems. Then, we investigated the performance of the DE approach for the large-scale stockpiles blending problem. We evaluated the proposed approach for real-data instances. The results show that the DE approach obtains in all cases significantly better results than the results of real-world situations. Next step, we are interesting to improve the performance of the approach and investigate other algorithms on the large-scale stockpile blending problem.

\section{Acknowledgements}
This research has been supported by the SA Government through the PRIF RCP Industry Consortium

\bibliographystyle{abbrv}
\bibliography{main}

\end{document}